\documentclass[10pt,conference]{IEEEtran}
\IEEEoverridecommandlockouts
\usepackage{cite}
\usepackage{amsmath,amssymb,amsfonts}
\usepackage{algorithmic}
\usepackage{graphicx}
\usepackage{textcomp}
\usepackage{xcolor}
\usepackage{booktabs}
\usepackage{mathtools}
\usepackage[english]{babel}
\usepackage{multirow}
\usepackage{subfig}
\usepackage{url}
\usepackage{hyperref}
\usepackage[justification=centering]{caption}

\newcommand{\comment}[1]{}

\def\BibTeX{{\rm B\kern-.05em{\sc i\kern-.025em b}\kern-.08em
    T\kern-.1667em\lower.7ex\hbox{E}\kern-.125emX}}

\AtBeginDocument{
  \catcode`_=12
  \begingroup\lccode`~=`_
  \lowercase{\endgroup\let~}\sb
  \mathcode`_="8000
}

\begin{document}

\title{"What makes my queries slow?": \\ Subgroup Discovery for SQL Workload Analysis}

%\author{Anonymous}

\author{
Youcef Remil\textsuperscript{1,2,3},
Anes Bendimerad\textsuperscript{2},
Romain Mathonat\textsuperscript{2},
Philippe Chaleat\textsuperscript{2},
Mehdi Kaytoue\textsuperscript{1,2}
\\\\
\textsuperscript{1}  Univ Lyon, INSA Lyon, CNRS, LIRIS UMR 5205, F-69621, Lyon, France
\\
\textsuperscript{2}  Infologic, 99 avenue de Lyon, 26500 Bourg-L{\`{e}}s-Valence, France
\\
\textsuperscript{3} \texttt{yre@infologic.fr}
}

\maketitle

\begin{abstract}

Among daily tasks of database administrators (DBAs), the analysis of query workloads to identify schema issues and improving performances is crucial. Although DBAs can easily pinpoint queries repeatedly causing performance issues, it remains challenging to automatically identify subsets of queries that share some properties only (a pattern) and simultaneously foster some target measures, such as execution time. Patterns are defined on combinations of query clauses, environment variables, database alerts and metrics and help answer questions like \textit{what makes SQL queries slow}? \textit{What makes I/O communications high?} Automatically discovering these patterns in a huge search space and providing them as hypotheses for helping to localize issues and root-causes is important in the context of explainable AI. To tackle it, we introduce an original approach rooted on \textit{Subgroup Discovery}. We show how to instantiate and develop this generic data-mining framework to identify potential causes of SQL workloads issues. We believe that such data-mining technique is not trivial to apply for DBAs. As such, we also provide a visualization tool for interactive knowledge discovery. We analyse a one week workload from hundreds of databases from our company, make both the dataset and source code available, and experimentally show that insightful hypotheses can be discovered.
\end{abstract}


\section{Introduction}\label{sec:introduction}

\begin{figure*}[t]
\centering
\begin{tabular}{cc}
\includegraphics[width=0.95\textwidth]{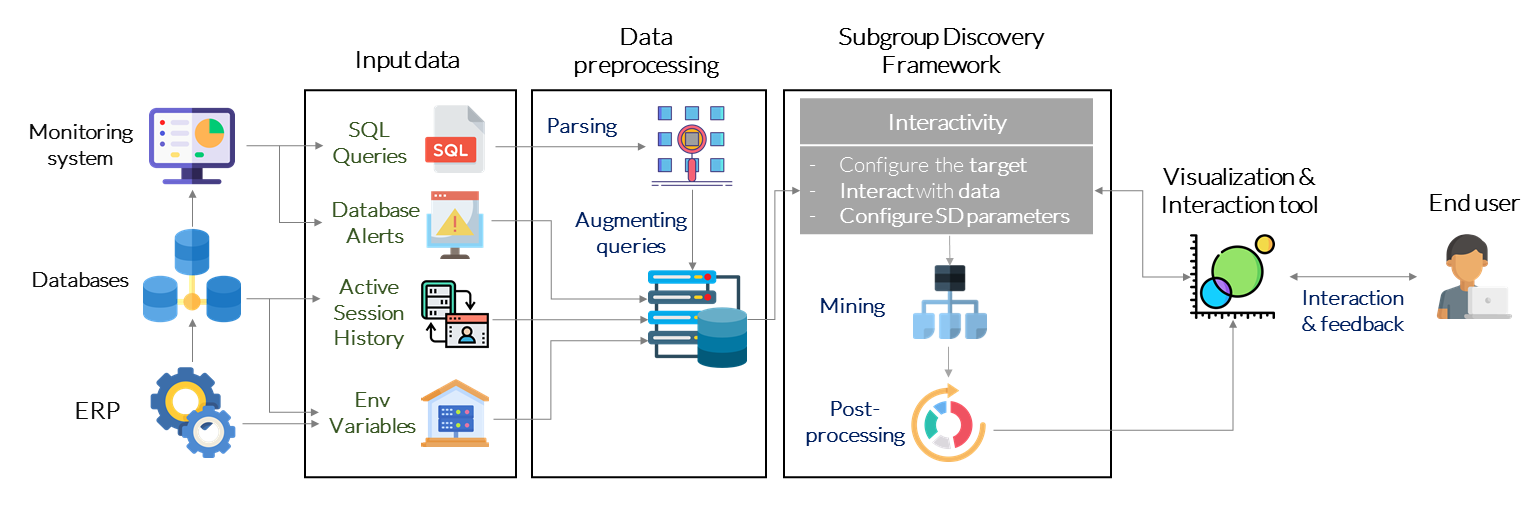}
\end{tabular}
\caption{\label{fig:overview} Overview of our Subgroup Discovery framework for SQL workload analysis.}
\end{figure*}

%According to the latest forecast by Gartner, Worldwide IT spending is projected to total \$3.9 trillion in 2021, an increase of 6.2\% from 2020~\cite{website:gartnerItForecast}. This increase is due to the high importance of Software industry and IT in different fields of our life. But also, an unprecedented speed of digital transformation has been implied by the 2020 Pandemic to satisfy remote working, education, and new social norms.  The increasing scale and complexity of this field poses significant challenges to engineers on efficiently maintaining high availability of software. Despite the high dedicated efforts put by these practitioners, software systems still encounter many incidents and outages that often cause serious economic loss. For example, the one-hour downtime of Amazon.com on Prime Day in 2018 (its biggest sale event of the year) caused the loss of up to \$100 million~\cite{website:amazonOutage}. In this context, the term AIOps (Artificial Intelligence for ITOps) has been introduced by Gartner~\cite{website:gartnerAIOps} to effectively address these challenges with AI. AIOps gathers different methods that use Artificial Intelligence techniques and~/~or Big data technologies to empower IT Operations, such as Predictive Maintenance and Software Analytics. 

It is indisputable that data has become a crucial part of software and IT platforms. This makes the database and its management system a critical component. Thus, researchers and engineers have spent a significant effort to make the interaction with data as reliable and efficient as possible. A data-driven strategy based on query workload analysis has proven its efficiency to address a large variety of related problems. These methods automatically analyze the set of logs and queries run on the database to perform tasks such as index recommendation~\cite{chaudhuri2003primitives,chaudhuri2002compressing}, query recommendation~\cite{akbarnejad2010sql,aligon2014similarity}, anti-pattern detection~\cite{arzamasova2017cleaning,eessaar2015query,chen2014detecting}, modeling user and application behavior~\cite{DBLP:conf/sigmod/TranMP15,DBLP:journals/tse/YuCHL92}. The usability of these data on such tasks strongly depends on their representation. That is why several methods have been proposed to transform the data into simplified forms before performing the main task, for example, workload compression~\cite{chaudhuri2002compressing}, efficient parsing~\cite{kul2018similarity} or embedding~\cite{jain2018query2vec} of SQL queries. Then, a myriad of Machine Learning methods have been evaluated on different workload analysis tasks. For example, clustering approaches have been exploited to delineate hot spots of user interests~\cite{arzamasova2019usefulness}, to summarize workloads~\cite{DBLP:journals/pvldb/YangPS09}, to identify insider threats~\cite{kul2016ettu}. NLP techniques have been used to embed SQL queries in vector representations that are guided by the target application~\cite{jain2018query2vec}. In~\cite{DBLP:conf/sigmod/ZolaktafMP20}, a neural network approach is proposed to help end-users and administrators compose SQL queries.

In this paper, we address a novel workload analysis problem that can uncover many kinds of tasks. We aim to design a method that efficiently brings answers to the generic question: \textit{how to characterize SQL queries that foster some properties of interest?} A concrete example of such question is: \textit{what makes queries slow?} Here the goal is to identify the characteristics and the context in which the execution time of queries is large. An example of results for such question is: 
$$\text{Table = X} \wedge \text{Where attribute = Y} \longrightarrow \text{high execution time}$$ 
Answering this question can be extremely useful for performance optimization problems. Similarly, several other questions of this kind may occur: how to characterize queries that over-consume the I/O communication? In which context SQL queries significantly increase concurrency issues? etc. To address this problem, we propose a unified and powerful framework rooted in the Subgroup Discovery approach. Subgroup Discovery~\cite{DBLP:books/mit/fayyadPSU96/Klosgen96,DBLP:journals/widm/Atzmueller15} is a data mining task that aims, among many other possibilities, to identify patterns describing parts in a dataset where the distribution of the target variable significantly deviates from the ``norm'', i.e. from its distribution in the whole dataset. Typically, the discovered subgroups are easily interpreted by the experts. Coming back to our previous example, the discovered subgroup consists of all the queries that verify the constraint ``$\text{Table = X} \wedge \text{Where attribute = Y}$". This subgroup is interesting because its average execution time is significantly greater than expected. 

Subgroup Discovery has proven its efficiency in different fields such as physics~\cite{goldsmith2017uncovering}, education~\cite{DBLP:journals/ijdsa/HelalLLEDM19} and neuro-science \cite{DBLP:journals/ploscb/LiconBSMFBGRPKB19}. To the best of our knowledge, our work is the first to exploit this approach to address a generic workload analysis problem.  The efficiency of such method is challenging as it strongly depends on several complex criteria: (1) the data needs to be introduced to the algorithm in the right format, (2) a relevant pattern language needs to be defined (the language used to select subsets of queries), (3) we need to choose the right function to measure the interestingness of a subgroup w.r.t. the target problem, (4) resulting subgroups need to be interpretable and their interactive mining needs to be enabled. 

\smallbreak
\noindent \textbf{Contributions.} We introduce an efficient Subgroup Discovery framework that meets all the aforementioned criteria for Workload Analysis. We first propose a data pre-processing step to prepare the SQL workload. We parse queries to extract important attributes (e.g., tables, fields, operations). We have extended the Mozilla parser~\cite{website:mozillaParser} with new features to extract all the information we need from the queries\footnote{Code and datasets available on \url{https://github.com/RemilYoucef/sd-4sql}} (e.g., handling alias and nested queries). Moreover, we augment queries with other relevant information: execution time, performance metrics of the DBMS, environment variables of the system, and anomaly alerts guided by expert knowledge. Then, we define a suitable pattern language, and integrate a diverse set of interestingness measures whose choice can be directed by the target application. We provide exact and heuristic algorithms to identify subgroups of interest. Furthermore, we integrate a visual tool that enables the user to interact with the algorithm, and iteratively learn from the provided results. The whole process of the proposed approach is summarized in Figure~\ref{fig:overview}.

\smallbreak
\noindent\textbf{Outline.} The remainder of the paper is organized as follows. Sec~\ref{sec:background} presents the raw data, the pre-processing strategy, as well as an informal description of the studied problem. Sec~\ref{sec:discoveringDiscSQL} formally defines the problem settings, introduces the interestingness measures in Sec~\ref{subsec:measures} and algorithms in Sec~\ref{subsec:algorithms} and Sec~\ref{subsec:postprocessing}. Then, Sec~\ref{subsec:visualMining} presents our interactive visual tool that enables the user to easily annotate data and design the target task. A thorough empirical study is detailed in Sec~\ref{sec:evaluation} to quantitatively and qualitatively evaluate the proposed approach. Sec~\ref{sec:relatedwork} presents related work before we conclude and present future directions. %Finally, we conclude in Sec~\ref{sec:conclusion}.

\section{Methodology}\label{sec:background}

We conduct our analysis on a workload $W$ of 150K unparametrized SQL statements gathered from more than 400 databases supervised in our company, sharing almost the same database schema. We efficiently parse queries to extract tables and attributes for each type of SQL clause. Queries are then augmented with several database metrics and supervision alerts, resulting in thousands of properties helping in contextualizing the subgroup discovery.

\begin{table*}
  \caption{Dataset features.}
  \label{tab: featsdata}
  \centering
  \scalebox{1}{
  \begin{tabular}{c|c|c|c|c|c}
    \toprule
    \textbf{Query properties} & \multicolumn{2}{c|}{\textbf{Environment variables}} & \textbf{Alerts} &\multicolumn{2}{c}{\textbf{Oracle ASH}}\\
    \midrule
    query & serverName       & dbMemory         & manyActiveSessions         & application &   concurrence \\
    day   & declination      & sgaMax           & blockedSessions       & configuration  &  network \\
    hour  & softwareVersion  & dbProcesses      & poolAlmostFull    & administrative & cpu \\
    time  & codeVersion      & jdbcMin          & anomalyASH   & systemI/O & userI/O \\
    nrows & dbVersion        & jdbcMax          &                       & queuing & scheduler \\
    &                        & dbCursorsMax     &                       & commit & \\
    \bottomrule
  \end{tabular} }
\end{table*}

\subsection{Raw Data}\label{sec:backgroundData}

\noindent\textbf{SQL queries.} We define the workload as a set $W = \{q_1, ..., q_n\}$ where each query $q \in W$ is a \textcolor{blue}{\texttt{SELECT}} statement that contains (1) the SQL text $q\textsubscript{text}$, (2) the query execution time $q\textsubscript{time}$ and (3) the number of rows returned $q\textsubscript{nrows}$.       

\smallbreak
\noindent \textbf{Environment features.} Each database is queried by an application, an Enterprise Resource Planning software (ERP) that we develop in the company. As such, the application identifier (\textit{serverName)} and its major and minor versions are considered (\textit{softwareVersion}, \textit{codeVersion}). The database properties are its vendor/version (\textit{dbVersion)}, its schema among $6$ main families (\textit{declination}),
the size of the database server memory (\textit{dbMemory}), the maximum database memory usage (\textit{sgaMax}), the maximum number of processes (\textit{dbProcesses}), the minimum and the maximum size of pool (\textit{jdbcMin}, \textit{jdbcMax}), the limit on the number of cursors per database session (\textit{dbCursorMax}).

\smallbreak
\noindent \textbf{Active Session History.} 
Active Session History (ASH)~\cite{website:ashOracle} was introduced in Oracle 10g, and then in other database systems such as PostgreSQL~\cite{website:ashPostgres}. An active session is a database session  waiting for some resource such as CPU, System I/O or Network. ASH provides the number of sessions waiting for each category of resource, per interval of time. It gives a temporal distribution that can be of high interest for diagnostics and tuning. 
For example, one may use this data to identify queries that unexpectedly over-consume network as they generate many network waiting sessions.

\smallbreak
\noindent \textbf{Alerts.} 
Our monitoring system triggers rule-based alerts when anomalies are observed on the database environment. For this analysis, we considered $4$ alerts: (1) when the number of active sessions is unusually large (\textit{manyActiveSessions}), (2) when some sessions remain blocked during a significant time (\textit{blockedSessions}), (3) when the size of the pool is close to its maximum limit (\textit{poolAlmostFull}), and (4) when there is an anomaly in the distribution of ASH (\textit{anomalyASH}), e.g., an increase in the proportion of sessions waiting for network or systemI/O. We have augmented the queries with the alerts that co-occur with their execution. Each alert has four levels: \textit{Info}, \textit{Alarm}, \textit{Critical} and \textit{blocking}.

\smallbreak
These features have been chosen with our DBAs. Our methodology is totally flexible and any numerical and categorical property can be considered as well. As related in Section \ref{sec:relatedwork}, it should be noticed that no prior work has invested such a combination of high dimensional features along with a very expressive representation of SQL queries that we present now.

\subsection{Query transformation}
\label{preprocessing}

\comment{\begin{table}
  \caption{SQL Parsing Keywords}
  \label{tab:keywords}
  \scalebox{1.05}{
  \begin{tabular}{c|c|c}
    \toprule
    \textbf{\textit{Clause name}} & \textbf{\textit{Start Keyword}} &\textbf{\textit{End Keyword}}\\
    \midrule
    Projections & \textcolor{blue}{\texttt{SELECT}} & \textcolor{blue}{\texttt{FROM}} \\ \midrule
    Tables & \textcolor{blue}{\texttt{FROM}} & \begin{tabular}[c]{@{}c@{}}
    \textcolor{blue}{\texttt{WHERE}}, \textcolor{blue}{\texttt{GROUP BY}}, \\ \textcolor{blue}{\texttt{ORDER BY}}, End of query
    \end{tabular} \\ \midrule
    Where attributes & \textcolor{blue}{\texttt{WHERE}} & \begin{tabular}[c]{@{}c@{}}
    \textcolor{blue}{\texttt{GROUP BY}}, \textcolor{blue}{\texttt{ORDER BY}}, \\ End of query
    \end{tabular} \\ \midrule
    Group by attributes & \textcolor{blue}{\texttt{GROUP BY}} & \begin{tabular}[c]{@{}c@{}}
    \textcolor{blue}{\texttt{ORDER BY}}, \textcolor{blue}{\texttt{HAVING}}, \\ End of query
    \end{tabular} \\ \midrule
    HAVING attributes & \textcolor{blue}{\texttt{HAVING}} & \begin{tabular}[c]{@{}c@{}}
    \textcolor{blue}{\texttt{ORDER BY}}, End of query
    \end{tabular} \\ \midrule
    Order by attributes & \textcolor{blue}{\texttt{ORDER BY}} &  End of query \\ \midrule
    Function attributes & \multicolumn{2}{c}{\begin{tabular}[c]{@{}c@{}} Tokens with 
    \textcolor{blue}{\texttt{AVG}}, \textcolor{blue}{\texttt{SUM}}, \textcolor{blue}{\texttt{COUNT}},
    \textcolor{blue}{\texttt{MIN}}, \textcolor{blue}{\texttt{MAX}}
    \end{tabular}} \\ 
    \bottomrule
  \end{tabular} }
\end{table}
}

A common preprocessing is to decompose, parse and tokenize SQL queries $q\textsubscript{text}$ to form a numerical vector where dimensions count the usage of data tables and attributes~\cite{aouiche2006clustering,aligon2014similarity,akbarnejad2010sql}. We have used the readily available Mozilla parser \cite{website:mozillaParser} that provides an SQL syntactic tree in XML that we then parse. We normalize the case sensitivity and remove irrelevant terms such as constants and logical operators. Tokens are then associated with the clauses they belong to, by adding to the token-name a prefix that indicates the clause in which each token appears. For instance in Figure~\ref{fig:exampleparsing}, the table \texttt{model} appears in the \textcolor{blue}{\texttt{FROM}} clause of the query, so its token will be \texttt{\textcolor{blue}{FROM}_model}.  For each token, we provide the number of time it appears in each SQL clause.

Then, we extended the Mozilla parser in two ways. Firstly we needed to consider not only SQL queries but also Hibernate queries used in our ERP as the ORM layer: we added the reserved keywords for hibernate queries such as \textcolor{blue}{\texttt{JOIN FETCH}} in the original parser. Second, and most importantly, unlike several existing parsers~\cite{aouiche2006clustering,aligon2014similarity,akbarnejad2010sql}, our parser can handle nested queries while preserving all the structure of the query. Furthermore, substitutes for temporary table names known as \textit{aliases} are not removed as also done in many parsers \cite{makiyama2015text,akbarnejad2010sql}. Aliases are rather used to figure out for example, to which table belongs a column in a \textcolor{blue}{\texttt{SELECT}} clause or a predicate in \textcolor{blue}{\texttt{WHERE}} or \textcolor{blue}{\texttt{GROUP BY}} clause. The interest in keeping and using alias appears when the queries are nested or the query contains a join clause, or involves several tables.  In this way, as the example in Figure~\ref{fig:exampleparsing} shows, if two columns of different tables have the same name, they will be encoded differently unlike the approach proposed by \cite{makiyama2015text}. Inspired by the work of \cite{deep2020comprehensive}, we also consider the function calls present in the SQL statement, as an independent clause. The source code is available as mentionned in the introduction.

Finally, it is noteworthy that we do not group semantically equivalent queries under a canonical form as done in \cite{kul2018similarity}: the way a query is written can impact its execution plan, thus, execution time.

\begin{figure}
    \centering
\scalebox{0.75}{
  \begin{tabular}{l|l}
    \toprule
    \multicolumn{2}{c}{\textbf{Raw SQL query}}   \\
    \midrule
    \multicolumn{2}{l}{\begin{tabular}[c]{@{}l@{}}
    \texttt{\textcolor{blue}{SELECT} m.ik}\\ 
    \texttt{\textcolor{blue}{FROM} model \textcolor{blue}{AS} m}\\
    \texttt{\textcolor{blue}{JOIN} prod \textcolor{blue}{AS} p}\\
    \texttt{\textcolor{blue}{WHERE} m.ik = p.ik}\\
    \hspace{2mm} \texttt{\textcolor{blue}{AND} m.uex = p1}\\
    \hspace{2mm} \texttt{\textcolor{blue}{AND} (m.uex \textcolor{blue}{in} collection0}\\
    \hspace{10mm} \texttt{\textcolor{blue}{OR} m.ik \textcolor{blue}{in} collection1)}\\
    \hspace{2mm} \texttt{\textcolor{blue}{AND} (m.dossierinfo = p3}\\
    \texttt{\textcolor{blue}{GROUP BY} m.ik}\\
    \texttt{\textcolor{blue}{HAVING} (\textcolor{blue}{COUNT}(\textcolor{blue}{DISTINCT} p.ik) = p2)}\\
    \texttt{\textcolor{blue}{AND} (\textcolor{blue}{SUM}(m.nbembal) = \textcolor{blue}{MAX} (p.nbembal))}
    \end{tabular}} \\
    \midrule
    \textbf{Our parsing result} & \textbf{Parsing result of} \cite{makiyama2015text} \\  
    \midrule
    \texttt{\textcolor{blue}{SELECT}_model.ik} $\longrightarrow$ 1 & \texttt{\textcolor{blue}{SELECT}_ik} $\longrightarrow$ 1\\ \midrule
     \texttt{\textcolor{blue}{FROM}_model} $\longrightarrow$ 1 & \texttt{\textcolor{blue}{FROM}_model} $\longrightarrow$ 1 \\
     \texttt{\textcolor{red}{JOIN}_prod} $\longrightarrow$ 1 & \texttt{\textcolor{red}{FROM}_prod} $\longrightarrow$ 1 \\ \midrule
     \texttt{\textcolor{blue}{WHERE}_\textcolor{red}{model.ik}} $\longrightarrow$ \textcolor{red}{3} & \texttt{\textcolor{blue}{WHERE}_\textcolor{red}{ik}} $\longrightarrow$ \textcolor{red}{4} \\
     \texttt{\textcolor{blue}{WHERE}_model.uex} $\longrightarrow$ 1 & \texttt{\textcolor{blue}{WHERE}_uex} $\longrightarrow$ 1 \\
     \texttt{\textcolor{blue}{WHERE}_model.dossierinfo} $\longrightarrow$ 1 & \texttt{\textcolor{blue}{WHERE}_dossierinfo}
     $\longrightarrow$ 1 \\
     \texttt{\textcolor{blue}{WHERE}_\textcolor{red}{prod.ik}} $\longrightarrow$ \textcolor{red}{1} &  \\ \midrule
     \texttt{\textcolor{blue}{GROUPBY}_{model.ik}} $\longrightarrow$ {1} & \texttt{\textcolor{blue}{GROUPBY}_{ik}} $\longrightarrow$ {1} \\ \midrule
     \texttt{\textcolor{blue}{HAVING}_{prod.ik}} $\longrightarrow$ {1} & \texttt{\textcolor{blue}{HAVING}_{ik}} $\longrightarrow$ {1} \\ 
     \texttt{\textcolor{blue}{HAVING}_{\textcolor{red}{model.nbembal}}} $\longrightarrow$ \textcolor{red}{1} & \texttt{\textcolor{blue}{HAVING}_{\textcolor{red}{nbembal}}} $\longrightarrow$ \textcolor{red}{2} \\
     \texttt{\textcolor{blue}{HAVING}_{\textcolor{red}{prod.nbembal}}} $\longrightarrow$ \textcolor{red}{1} & \\ \midrule
     
     \texttt{\textcolor{blue}{COUNT}_{prod.ik}} $\longrightarrow$ {1} & \\
     \texttt{\textcolor{blue}{SUM}_{model.nbembal}} $\longrightarrow$ {1} & \\
     \texttt{\textcolor{blue}{MAX}_{prod.nbembal}} $\longrightarrow$ {1} & \\
    \bottomrule
  \end{tabular}}
  \caption{Example of parsing an SQL query.}
  \label{fig:exampleparsing}
\end{figure}

\subsection{Data Model}
\begin{table*}
    \centering
    \scalebox{0.83}{
  \begin{tabular}{c|c|c|c|c|c|c|c|c|c|c|c|c}
    \toprule
     \multirow{2}{*}{$O$ } & \multicolumn{2}{c|}{\textcolor{blue}{\texttt{\textbf{FROM}}}} & \multicolumn{3}{c|}{\textcolor{blue}{\texttt{\textbf{WHERE}}}} & \multicolumn{2}{c|}{\textbf{ENV features}} & \textbf{Alerts} & \textbf{ASH} & $q\textsubscript{nrows}$ & \multicolumn{2}{c}{$q\textsubscript{time}$} \\ 
     \cmidrule(l){2-13}
     & \begin{tabular}[c]{@{}c@{}}$a_1$ \\\texttt{Verrou} \end{tabular} & 
       \begin{tabular}[c]{@{}c@{}}$a_2$ \\\texttt{Cumulof} \end{tabular} &
       \begin{tabular}[c]{@{}c@{}}$a_3$ \\\texttt{Verrou.ik} \end{tabular} &
       \begin{tabular}[c]{@{}c@{}}$a_4$ \\\texttt{Verrou.date} \end{tabular} &
       \begin{tabular}[c]{@{}c@{}}$a_5$ \\\texttt{Cumulof.ik} \end{tabular} &
       \begin{tabular}[c]{@{}c@{}}$a_6$ \\ Soft version \end{tabular} &
       \begin{tabular}[c]{@{}c@{}}$a_7$ \\ Server name \end{tabular} &
       \begin{tabular}[c]{@{}c@{}}$a_8$ \\ manyActiveSessions \end{tabular} &
       \begin{tabular}[c]{@{}c@{}}$a_9$ \\ Concurrency \end{tabular} &
       \begin{tabular}[c]{@{}c@{}}$a_{10}$ \\ nrows\end{tabular} &
       \begin{tabular}[c]{@{}c@{}}$a_{11}$ \\ time\end{tabular} &
       \begin{tabular}[c]{@{}c@{}}$a_{12}$ \\ slow\end{tabular} \\
    \toprule 
    $o_1$     & 1 & 0 & 1 & 0 & 0 & v2 & LYN & Alarm    & 22  & 10  & 2.15 & 0  \\ \hline
    $o_2$     & 1 & 0 & 1 & 1 & 0 & v1 & BLV & Critical & 3  & 1   & 15.81 & 1\\ \hline
    $o_3$     & 0 & 1 & 0 & 0 & 1 & v1 & BLV & Critical & 15 & 27  & 1.14  & 0\\ \hline
    $o_4$     & 1 & 1 & 0 & 1 & 1 & v2 & LYN & Alarm    & 31  & 12  & 10.87 & 1 \\ \hline
    $o_5$     & 1 & 1 & 1 & 0 & 1 & v3 & LYN & Alarm    & 11 & 25  & 2.1 & 0  \\ \hline
    $o_6$     & 1 & 0 & 1 & 2 & 0 & v3 & LYN & Critical & 6  & 100 & 17.93 & 1 \\ \hline
    $o_7$     & 1 & 1 & 1 & 1 & 1 & v2 & LYN & Info     & 27 & 1   & 15.8 & 1 \\ \hline
    $o_8$     & 0 & 1 & 0 & 0 & 1 & v2 & BLV & Alarm    & 9 & 37  & 9.95 & 0 \\ \hline
    $o_9$     & 1 & 0 & 1 & 0 & 0 & v3 & BLV & Critical & 10 & 112 & 8.95 & 0 \\ \hline
    $o_{10}$  & 0 & 1 & 0 & 0 & 1 & v2 & BLV & Alarm    & 7 & 1   & 14.7 & 1 \\ \hline
    $o_{11}$  & 0 & 1 & 0 & 0 & 0 & v2 & LYN & Info     & 25 & 16  & 1.0 & 0  \\ 
    \bottomrule
  \end{tabular}}
  \caption{Toy Example of a dataset $(O,A)$.}
  \label{fig:toyexample}
\end{table*}

We unify the different data sources into a dataset defined by a pair $(O,A)$, where $O=\{o_i\}_{1\leq i \leq n}$ is a set of objects that refer to the queries, and $A=(a_j)_{1 \leq j \leq m}$ is a vector of attributes. Each attribute $a : O \longrightarrow dom(a)$ is a function that maps queries to values in its domain $dom(a)$. Consequently, $a(o)$ denotes the value of the attribute $a$ for the object $o$. $dom(a)$ is given by $\mathbb{R}$ if $a$ is \textit{numerical}, by a finite set of categories $C_i$ if $a$ is \textit{nominal (categorical)}, or by $\{0,1\}$ if $a$ is \textit{Boolean}. A nominal attributes with a total ordering of its values is called an \textit{ordinal nominal} attribute. These notations are illustrated in Table~\ref{fig:toyexample} with a dataset of 11 objects $O = \{o_1, ..., o_{11} \}$ referring to queries described by 12 attributes. \textit{Server name} is nominal and has two possible values: \textit{LYN} and \textit{BLV}. \textit{manyActiveSessions}, referring to an alert, is an ordinal attribute with 3 levels \{\textit{Info}, \textit{Alarm}, \textit{Critical}\}. \textit{time} is numerical and gives the execution time that a query takes. \textit{slow} is binary, it equals $1$ when the query time exceeds $10$ seconds, $0$ otherwise. As mentioned in Sec~\ref{preprocessing}, when parsing a query, we keep the count of each token associated with each clause, thus, each token is a numerical attribute $a_j \in A$ such that : $dom (a_j) = \mathbb{N} \subset \mathbb{R}$. For example,  \texttt{Verrou.data} is numerical and represents the number of times this attribute appears in the \textcolor{blue}{\texttt{WHERE}} clause for each query, e.g., $a_4(o_6) = 2$ means that for the 6-th query in the dataset the attribute \texttt{Verrou.date} appears twice in the \textcolor{blue}{\texttt{WHERE}} clause.

\subsection{Characterizing Discriminant Queries}
The goal of subgroup discovery is to find subsets of objects that are statistically \textit{the most interesting} with respect to a property of interest i.e., the \textit{target}. For example, we seek to characterize queries that have large execution times, i.e., \textit{slow} queries. Thus, the \textit{target concept} can be the binary attribute \textit{slow}, and we will identify interpretable descriptions of subgroups that maximize the proportion of queries having $slow=1$. In Table~\ref{fig:toyexample}, a discriminant subgroup can be defined by queries that include the attribute \texttt{\textcolor{blue}{WHERE}_Verrou.date}. Indeed, 100\% of these queries are \textit{slow}, while only 45\% of overall queries are \textit{slow}. We refer to this subgroup by its description : ``$\text{\texttt{\textcolor{blue}{WHERE}_Verrou.date}} > 0$". Another interesting example consists in queries that correspond to the following description: ``$\text{\texttt{\textcolor{blue}{WHERE}_Cumulof.ik}} = 1 \wedge \text{\texttt{Soft. version} = \text{v2}}$", with a proportion of 75\% of \textit{slow} queries. Given a large number of attributes, we end up with a very huge set of possible conjunctive combinations. Therefore, it becomes challenging to identify those descriptions that are the most significantly discriminant. This is where an automatic Subgroup Discovery approach can be extremely helpful. Such approach usually identifies interesting results by performing a deep search through the set of candidates hypotheses and scores each of them with a function that assesses their interestingness. In addition to this purely automatic approach, human expertise can be useful in guiding the search, since subgroup discovery involves an iterative and interactive process. The first subgroup (``$\text{\texttt{\textcolor{blue}{WHERE}_Verrou.date}} > 0$") can inform experts that an index is probably missing on the attribute \texttt{Verrou.date}. The previous examples use a binary attribute (\textit{slow}) as the target concept, but subgroup discovery can also employ numerical targets or even complex models over multiple targets. In the following, consider that the target concept is the numerical attribute \textit{time}. The subgroup defined as ``$\text{\texttt{\textcolor{blue}{FROM}_Verrou}} > 0 \wedge \text{manyActiveSessions = \textit{Critical}}$" is statistically interesting, because its average \textit{time} of 14.22s is relatively large compared to the average over the whole dataset estimated by 9.12s. The last example would not have been impressive if we opted for the binary target \textit{slow}, because this subgroup is characterized by $3$ queries, one of which is barely less than 10s. It is also noteworthy that the size of subgroups is often taken into account to assess their quality. In fact, we are generally interested by discriminant subgroups that cover a large number of queries, as statistically more significant. 

\section {Discriminant Pattern Discovery in Workloads \label{sec:discoveringDiscSQL}}
\subsection{Introduction to Subgroup Discovery}

\noindent \textbf{Descriptive attributes and target.} One needs to specify a target attribute $t \in A$ that is suitable for the target application. For example, if we want to characterize slow queries, the target attribute $t$ will be the execution time ($q_{time}$). In this paper, we consider both cases where $t$ can be Boolean or numerical. Another important question is: which attributes do we want to use to characterize interesting subgroups? These are called descriptive attributes and denoted $A_D \subseteq A \setminus \{ t\}$, and $|A_D|=m_D$. A pattern language $\mathcal{D}$ is then defined over descriptive attributes. A pattern $d \in \mathcal{D}$ is a constrained selector of subset of objects using their descriptive attribute values $A_D$. More precisely, the pattern language is defined as $\mathcal{D} = \bigtimes_{i=1}^{m_D} \mathcal{D}_i$ where $\mathcal{D}_i$ is a selector defined over $a_i \in A_D$, and given by the set of all possible intervals in $\mathbb{R}$ if $a_i$ is numerical, the set $\{C_i,\emptyset\} \cup \{\{c\} \mid c \in C_i\}$ if $a_i$ is categorical, or $\{\{0,1\},\{0\} ,\{1\}\}$ if $a_i$ is Boolean. A pattern $d \in \mathcal{D}$ is then given by a set of restrictions over each descriptive attribute (i.e. $d = (d_i)_{1 \leq i \leq m_D}$). 

\smallbreak
\noindent\textbf{Linking patterns and objects.} A pattern $d = (d_i)_{1 \leq i \leq m_D} $ is said to \emph{cover} an object $o \in O$ iff $\forall a_i \in A_D : a_i(o) \in d_i$. The set of all objects covered by a pattern $d$ is called the extent of $d$ and denoted $ext(d)=\{o \in O \mid d \textit{ covers } o\}$.
In Table~\ref{fig:toyexample}, consider the case where we have 3 descriptive attributes $A_D=\{$\texttt{\textcolor{blue}{FROM}_Verrou}, \texttt{\textcolor{blue}{FROM}_Cumulof}, \texttt{Server name}$\}$. An example of pattern is $d=($\texttt{\textcolor{blue}{FROM}_Verrou}~$\in \mathbb{N},$ \texttt{\textcolor{blue}{FROM}_Cumulof}~$\geq 1,$ \texttt{Server name}~=~LYN$)$. This pattern covers objects in which \texttt{\textcolor{blue}{FROM}_Cumulof} appears at least once in the query and the \texttt{Server name} is LYN. These objects are $ext(d)=\{o_4,o_5,o_7,o_{11}\}$.

\smallbreak
\noindent \textbf{Subgroup definition.} A subgroup is any subset of objects $s \subseteq O$ that can be selected using a pattern $d$ over descriptive attributes $A_D$. The set of all possible subgroups is denoted $\mathcal{S}=ext(\mathcal{D})=\{ext(d) \mid d \in \mathcal{D}\}$. In other terms, a subgroup is a set of objects that can be characterized with some restrictions of attributes, turning it interpretable to the user. 

\smallbreak
\noindent \textbf{Subgroup interestingness.} A measure $\phi: \mathcal{S} \to \mathbb{R}; s \mapsto \phi(s)$ is a mapping that evaluates the quality of a subgroup $s$ w.r.t. the property of interest. The greater is $\phi(s)$, the more interesting is $s$. The choice of $\phi$ depends on the target application. In Sec~\ref{subsec:measures}, we define several relevant measures that we have exploited to analyze SQL queries, and we explain how to choose the right measure according to the end user goal.

\smallbreak
\noindent \textbf{Problem statement.} Given a user specified parameter $k$, find the top-$k$ subgroups with the highest values of the interestingness measure $\phi$. Formally, find the subgroup set: 
%$$\mathcal{R}=\{ s \in \mathcal{S} \text{ such that: } |\{ s' \in \mathcal{S} \mid \phi(s')>\phi(s) \}|<k \}.$$
$$\mathcal{R}=\{ s \in \mathcal{S}  \mid rank(s) \leq k \},$$
where $rank(s)$ gives the rank of $s$ w.r.t. its score $\phi$, that is: $rank(s)=|\{ s' \in \mathcal{S} \mid \phi(s')>\phi(s) \}| +1 $.

\subsection{Measuring subgroup interestingness}\label{subsec:measures}
We present measures that can be used as $\phi$ to assess the quality of subgroups. It is generally agreed that interesting discriminant subgroups are those that maximize the deviation of the target attribute $t$ and whose size $|s|$ is sufficiently large. In fact, one prefers discriminant subgroups that have large sizes as they are deemed more significant, i.e., there existence in the dataset is less probable to be due to chance. Many of existing measures belong to the popular family of \texttt{Kl\"{o}sgen} functions~\cite{DBLP:books/mit/fayyadPSU96/Klosgen96} defined given the parameter $a \in [0,1]$:
$$
\texttt{Kl\"{o}sgen}_a(s)=sup(s)^a \cdot \left( \mu(s) - \mu(O) \right),
$$
where the support $sup(s)=\frac{|s|}{|O|}$ measures the proportion of objects from $O$ that belongs to $s$, and the target mean $\mu(s)=\sum_{o \in s} \frac{t(o)}{|S|}$ gives the average value of the target attribute $t$ in $s$. Thus, the higher is $sup(s)$, the higher is $\texttt{Kl\"{o}sgen}_a(s)$. But also, $\texttt{Kl\"{o}sgen}_a(s)$ is maximized when the  deviation of $\mu(s)$ regarding its overall mean $\mu(O)$ is maximized. The choice of $a$ affects the importance of $sup(s)$ on the final value of interestingness, and conducts to measures with different statistical interpretations. These measures are presented in what follows.

\smallbreak
\noindent \textbf{Average function $u$.} Also called unusualness~\cite{DBLP:journals/jmlr/SchefferW02}, it is the $\texttt{Kl\"{o}sgen}$ function with $a=0$, that is: $u(s)=\mu(s)-\mu(O)$. It can be used when we do not want to impact the score of subgroups by $sup(s)$. When used, this measure is generally combined with a threshold constraint on a minimum size of returned subgroups, to avoid retrieving very small ones. $u(s)$ provides a subgroup ordering that is identical to another popular measure: $\texttt{Lift}(s)=\frac{\mu(s)}{\mu(O)}$.

\smallbreak
\noindent \textbf{\texttt{WRAcc} measure~\cite{DBLP:journals/jmlr/LavracKFT04}.} It is one of the most popular measures in SD. It corresponds to the $\texttt{Kl\"{o}sgen}$ function with $a=1$: $$\texttt{WRAcc}(s)=sup(s) \cdot \left( \mu(s) - \mu(O) \right).$$ 
For the specific case when $t$ is binary, it can be written as: 
$$\texttt{WRAcc}(s)=\texttt{Pr}( o \in s \wedge t(o)=1) - \texttt{Pr}(o \in s) \cdot \texttt{Pr}(t(o)=1),$$
where $\texttt{Pr}$ is the probability of an event to happen. Theoretically, the more $s$ is statistically dependent of true target values ($t(o)=1$), the higher is $|WRAcc(s)|$.

\smallbreak
\noindent \textbf{\texttt{Mean-test}.} A drawback of \texttt{WRAcc} is that, in many tasks, it over-scores subgroups with large support despite their limited unusualness. For this reason, many methods have preferred to use the \texttt{Mean-test}, which is the $\texttt{Kl\"{o}sgen}$ function with $a=0.5$:
$$\texttt{Mean-test}(s)=\sqrt{sup(s)} \cdot \left( \mu(s) - \mu(O) \right).$$ 
From a statistical point of view, it was proven that this measure provides an equivalent ordering than the Binomial test~\cite{DBLP:books/daglib/0035669}.

\smallbreak
\noindent \textbf{\texttt{T-score}.} A limitation of \texttt{Kl\"{o}sgen} functions is that they do not optimize the dispersion of the target attribute in subgroups. This could lead to inconsistent statements, particularly when the dataset contains many outliers. In fact, $\mu(s)$ is sometimes not representative of target values in the subgroup $s$, if it contains few outliers with extreme values of $t$. To address this issue, one of the measures that incorporate the cohesion of the subgroup is the \texttt{T-score}~\cite{pieters2010subgroup}, defined as:
$$\texttt{T-score}(s)=\frac{\sqrt{sup(s)}}{\sigma(s)} \cdot \left( \mu(s) - \mu(O) \right),$$ 
where $\sigma(s)$ is the standard-deviation of target values $t$ in the subgroup $s$. The smaller is $\sigma(s)$, the more cohesive are values of $t$ in $s$, and thus the higher is $\texttt{T-score}(s)$. This measure reflects the significance of the deviation of target values in a subgroup using a Student’s t-test. However, one should avoid a direct statistical interpretation of the \texttt{T-score} if the target attribute is not normally distributed and the subgroup size is small, e.g., $|s| < 30$.

\smallbreak
\noindent \textbf{Median-based measures \texttt{q_{med}}.} Another way to reduce the impact of outliers on subgroup scores is to estimates values of $t$ in $s$ using its median $\texttt{med}(s)$ instead of its average $\mu(s)$ in the \texttt{Kl\"{o}sgen} function, as the median estimator is more robust to noise~\cite{DBLP:journals/datamine/LemmerichAP16}:
$$
\texttt{q_{med}}(s)= sup(s)^a \cdot \left( \texttt{med}(s) -  \texttt{med}(O) \right).
$$

\subsection{Algorithms}\label{subsec:algorithms}
Once the pattern language is defined and the subgroup interestingness is chosen, it remains to explore the search space in order to identify the top-$k$ subgroups. Computational complexity of SD problem is known to be prohibitive due to the huge size of the search space $|\mathcal{S}|$ that increases exponentially w.r.t. $|A_D|$. Many algorithms have been proposed to efficiently traverse the search space. Some of them provide exact results, others are heuristic but scale better. \cite{DBLP:conf/pkdd/Lemmerich018} proposed a Python implementation of the most popular SD algorithms. Since it does not support all the relevant measures for our case study, we have extended this framework to incorporate: the support, the \texttt{T-score}, as well as median-based measures \texttt{q_{med}}. We exploit two methods: (1) an exact algorithm based on a depth-first search, (2) a heuristic algorithm that uses beam-search.
%\footnote{\url{https://tinyurl.com/nhk6hfkt}}

\smallbreak
\noindent \textbf{Depth-first algorithm.} This approach exhaustively explores the search space $\mathcal{S}$ in a depth-first manner. After defining an order relation between patterns, the search space forms a lattice structure with the empty pattern as a supremum and the pattern containing all the selectors as infimum. Then, this lattice is explored in depth. We start from the empty pattern $d=\left(a_i \in dom(a_i) | a_i \in A_D \right)$, i.e., no restriction for any attribute. Then, a refinement operator is recursively applied on selectors of $d$, continuously making it more restrictive. Refinements can be operated by adding a symbolic attribute value, or adding a numerical attribute cut point. As the search space can be extremely large, a naive enumeration of subgroups fails. For this reason, the exact algorithm uses many techniques to optimize the exploration. Some anti-monotonic constraints are generally used, such as a minimum support $\delta$, i.e., if a pattern covers less than $\delta$ objects then this pattern is not refined anymore, as its refinement necessarily covers less than $\delta$ objects. Furthermore, tight optimistic estimates \texttt{TOE}~\cite{DBLP:journals/datamine/LemmerichAP16} are used. These functions allow to efficiently upper bound all the subgroup interestingness values in a whole branch of the search space. If the \texttt{TOE} of a branch is lower than the score of the top-$k$ already found subgroup, then the branch is pruned, as it does not contain any subgroup with a score higher than the already found top-$k$. Other optimization strategies are used as well. to~\cite{DBLP:journals/datamine/LemmerichAP16} for further details.

\smallbreak
\noindent \textbf{Beam-search algorithm.} Heuristic methods are deemed useful in many scenarios when the number of descriptive attributes is large, and thus exact methods become slow or infeasible. They try to find as good patterns as possible in a short time by evaluating only promising candidates. The most popular heuristic approach is Beam-search~\cite{DBLP:journals/ml/ClarkN89}. This approach performs a heuristic level-wise search over the pattern lattice. It requires to specify the width parameter $w \in \mathbb{N}$, which is the maximum number of patterns kept in each level of the lattice. It starts from the empty pattern $d=\left(a_i \in dom(a_i) | a_i \in A_D \right)$. Then, it recursively goes to the next level by refining patterns of the current level and selecting the top-$w$ refined patterns that maximize the interestingness. These top-$w$ patterns are then refined again to continue to a deeper level. At the end, the algorithm selects the top-$k$ subgroups among all the top-$w$ ones selected from each level.

\smallbreak
\subsection{Reducing information redundancy}\label{subsec:postprocessing}
The process of selecting interesting subgroups considering only the discriminative measure may result in strongly overlapping subgroups, that are distinct patterns covering almost the same objects. 
%These subgroups are potentially caused by the same hypothesis.
%Since our goal is to find the best-$k$ subgroup patterns that maximize an interestingness measure $\phi$, other more interesting subgroups will not be considered.
For instance in Table~\ref{fig:toyexample}, the two different patterns : $d_1 = (\text{\texttt{\textcolor{blue}{FROM}_Verrou}}~\geq 1)$ and $d_2 = (\text{\texttt{\textcolor{blue}{WHERE}_Verrou.ik}}~\geq 1)$ are highly correlated as they cover more or less the same objects. In order to provide interesting but diverse patterns, and to reduce information redundancy in the subgroup set $\mathcal{R}$, we propose two different solutions based on the \textit{Jaccard similarity}~\cite{niwattanakul2013using}. This metric measures the similarity of two subgroups patterns as a fraction between the intersection and the union of their extents. For the two example patterns $d_1$ and $d_2$: 
$$
sim(d_1, d_2) = J(d_1, d_2) = \frac{|ext(d_1) \cap ext(d_2)|}{|ext(d_1) \cup ext(d_2)|} = \frac{6}{7}
$$         

\smallbreak
\noindent \textbf{Greedy Selection.} The greedy approach constructs iteratively the non redundant subgroup set $\mathcal{R^\prime}$.  In each iteration, the best subgroup $s^{\star}$ in the initial subgroup set $\mathcal{R}$ is identified, and added to $\mathcal{R^\prime}$. Afterwards,
we remove from $\mathcal{R}$ all the subgroups whose similarity with $s^{\star}$ exceeds a specified threshold.
This process is repeated until $\mathcal{R}$ becomes empty. However, this technique requires the user to specify an appropriate threshold for the \textit{Jaccard similarity}. For a large enough threshold, one can still end up with overlapping subgroups. On the other side, a small threshold can lead to the suppression of interesting subgroups. Thus, this method is sensitive to the threshold which must be chosen empirically.     

\smallbreak
\noindent \textbf{Hierarchical Clustering.} To get a more complete and understandable overview of the resulting subgroups, agglomerative hierarchical clustering \cite{DBLP:conf/flairs/AtzmullerP08} is performed on the result set $\mathcal{R}$. At the bottom of the hierarchy, each subgroup forms a singleton cluster, and pairs of clusters are then merged as one moves up the hierarchy. This clustering is computed using the \textit{Jaccard dissimilarity} defined as $1 - \text{\textit{Jaccard similarity}}$. As a result, the hierarchical clustering produces a binary clustering tree or a \textit{dendrogram}. It represents a hierarchy of partitions, hence, it is possible to choose one partition by truncating the tree at a given level. Unlike the greedy approach, the user can specify how many non-redundant subgroup patterns she wants without having to specify a dissimilarity threshold.           

\subsection{Interactive SD for Workload Analysis}\label{subsec:visualMining}
In practice, an effective SD approach needs to be iterative and interactive, to make it possible to incorporate subjective criteria as well as human expertise. Indeed, the interestingness of subgroups strongly depends on the end user preferences and her prior knowledge about the data. This interactive process should efficiently allow the user to explore the region of her hypothesis space, and possibly improve the quality of the extracted pattern. In that process, the user sets the parameters of the approach including the measure and the algorithm to get a visualisation of the retrieved patterns. Afterwards, the user proceeds to the validation of these patterns and checks for the quality of the provided knowledge, while guiding the post processing phase to refine relevant patterns. Several subgroup discovery algorithms have been embedded into software tools such as KEPLER \cite{DBLP:conf/kdd/WrobelWSE96}, SubgroupMiner \cite{DBLP:conf/pkdd/KlosgenM02} and VIKAMINE \cite{DBLP:conf/pkdd/AtzmuellerL12} that provide a graphical interface allowing the user to choose viewing options and select appropriate parameters for her task. Different from those just mentioned, the graphical tool we provide enables the user to: (1) use visual filters to select the subset of the data she wants to mine with SD, and (2) visually constitute a binary \textit{target concept} that she aims to discriminate, by flexibly selecting data inside widgets such as scatter plots. Moreover, the tool allows for setting the desired parameters of the task and the visualization of the extracted patterns along with their associated statistics. Figure~\ref{fig:interacttool} shows the main window of this tool, which is described in Sec~\ref{subsec:interactive}.

\begin{table*}
  \caption{Datasets statistics.}
  \label{tab: statdata}
  \centering
  \renewcommand{\arraystretch}{0.5} 
  \scalebox{0.9}{
  \begin{tabular}{c|c|c|c|c|c|c|c|c|c|c}
    \toprule
    \textbf{Dataset} & 
    \textbf{Queries} & 
    \textbf{Features} & 
    \textbf{\textcolor{blue}{\texttt{FROM}} Tables} & 
    \textbf{\textcolor{blue}{\texttt{JOIN}} Tables} & 
    \textbf{Projections} & 
    \textbf{\textcolor{blue}{\texttt{WHERE}} atts} & 
    \textbf{\textcolor{blue}{\texttt{HAVING}} atts} & 
    \textbf{\textcolor{blue}{\texttt{GROUPBY}} atts} & 
    \textbf{\textcolor{blue}{\texttt{ORDERBY}} atts} & 
    \textbf{Sparsity} \\ 
    \midrule 
    All & 148796 & 8691 & 497 & 526 & 3740 & 3294 & 10 & 199 & 391 & 99.55\% \\
    \midrule
    D1 & 37149  & 4596 & 275 & 270 & 2036 & 1680 & 10 & 96 & 196 & 99.22\% \\
    \midrule
    D2 & 48823  & 246 & 1 & 1 & 86 & 85 & 2 & 21 & 11 & 84.27\% \\
    \midrule
    D3 & 3031  & 570 & 58 & 30 & 158 & 275 & 3 & 6 & 15 & 94.77\% \\
    \midrule
    D4 & 26735 & 3723 & 218 & 234 & 1658 & 1324 & 10 & 91 & 154 & 98.97\% \\
    \bottomrule
  \end{tabular} }
\end{table*}

\section{Evaluation}\label{sec:evaluation}
We report the experimental study that we conducted to evaluate the efficiency of our Subgroup Discovery approach. First, we validate that the proposed framework is able to characterize discriminant subgroups that are \textit{statistically} the most interesting w.r.t. different target problems. Then, we report through a quantitative analysis the execution time for each algorithm with different parameters including the number of patterns $k$ and rules depth. Finally, we present the different features provided by our visualization tool. Further study of measures and post-processing are differed in the supplementary materials due to lack of space. 

\smallbreak
\noindent \textbf{Experiments Setup.}  
Experiments are conducted on an SQL workload that contains hibernate queries run on our production-environment servers for a period of one week. Since the effective number of queries is extremely large, our monitoring system records only those whose execution time exceeds $5$ seconds.
We augment these queries with other relevant information described in Table~\ref{tab: featsdata}, as explained in Sec~\ref{sec:backgroundData}. 
This dataset contains 148,796 queries described by 8,691 features. Table~\ref{tab: statdata} displays the characteristics of the overall dataset.
Note that the dataset is extremely sparse, with only 0.45\% of non-zero values. The proposed framework extended the library Pysubgroup \cite{DBLP:conf/pkdd/Lemmerich018} to support more relevant measures, specifically for numerical targets (e.g., median-based measure). All the experiments presented were run on a single machine with (Intel(R) Core(TM) i5-10210U CPU @ 1.60GHz 2.11 GHz with 32GB RAM).

\subsection{Qualitative Analysis} 
\noindent \textbf{Use cases.} 
To show the ability of the proposed framework to perform several tasks with different goals, we use it to address the following diverse set of research questions:
\begin{itemize}
    \item \textbf{RQ1}: What makes queries very slow? 
    \item \textbf{RQ2}: In which context do queries present concurrency issues? 
    \item \textbf{RQ3}: How to characterize queries that co-occur the most with alerts of type \textit{blocked sessions}.
\end{itemize}

\smallbreak
\noindent \textbf{Methodology.} 
To properly conduct these experiments, we assess each use case in a specific context according to the industrial needs that we are facing. More precisely, each use case is carried out on a subset of data as follows: We evaluate \textbf{RQ1} on \textbf{D1}: queries that were executed on all sales servers with at least 100 users. Note that the sales declination constitutes 74.04\% of the data. \textbf{D2}: queries that invoke the \texttt{MVTREALISE} table, knowing that it is the most commonly queried table. \textbf{RQ2} is evaluated particularly on \textbf{D3}: the software version (V15\_2), since it presents exclusively many concurrency issues, compared to other different software versions. For this purpose, we consider a binary target which refers to a potential issue if there is at least an average of 5 concurrent processes over a period of 10 seconds during the execution of the query. Finally, we evaluate \textbf{RQ3} on \textbf{D4}: a specific set of servers on which we observe an abnormal raising of blocked session alerts. The characteristics of the 4 sub-datasets are provided in Table~\ref{tab: statdata}. For each of the studied scenarios, we choose the top-10 subgroups w.r.t. to the most adequate interestingness measure for the problem. The choice of the measure is made empirically by comparing the subgroups identified with each measure. We take the one that provides relevant but also interpretable patterns by performing a statistical distribution analysis and referring to human expertise. The resulting patterns are then processed to provide diverse and non-redundant ones, as described in Sec~\ref{subsec:postprocessing}. Results are given in Table~\ref{tab: sdresult} where for each subgroup pattern, we show its support, as well as its deviant quality compared to the dataset. For binary target problems, we compare the precision of each subgroup $s$ given by : $prec (s) = \frac{|o \in s \mid t(o) = 1|}{|s|}$ with the precision of the considered dataset.

\begin{table*}
  \caption{Subgroup Discovery Results.}
  \renewcommand{\arraystretch}{1} 
  \label{tab: sdresult}
  \centering
  \scalebox{0.84}{
  \begin{tabular}{c|c|c|l|c|c}
    \toprule
    \textbf{ID} & 
    \textbf{Target} & 
    \textbf{Measure} & 
    \multicolumn{1}{c|}{\textbf{Subgroup patterns}} & 
    \textbf{Size} &
    \multicolumn{1}{c}{\textbf{Quality}} \\ 
    \midrule 
    {D1} & \begin{tabular}[c]{@{}c@{}} \textbf{time} \\ (Numerical) \end{tabular} & Median & \begin{tabular}[l]{@{}l@{}} $(s_1):\text{\texttt{\textcolor{blue}{WHERE}_stocks.gestion.modele.lot.prod.ref.auditinfo.etat}} > 0$ \\ $(s_2):\text{\texttt{\textcolor{blue}{FROM}_ventes.cumuls.modele.cumulmultiple}} > 0$\\ $(s_3):\text{\texttt{\textcolor{blue}{WHERE}_ventes.cumuls.modele.cumulmultiple.valzvcli\textbf{X}}} > 0$ \\ $(s_4):\text{\texttt{\textcolor{blue}{WHERE}_.ventes.cumuls.modele.cumulmultiple.valzvart\textbf{X}}} > 0$\end{tabular} & \begin{tabular}[c]{@{}c@{}}8\\ 451\\ 45 \\ 45\end{tabular} & \begin{tabular}[l]{@{}l@{}}$161 \times med\_dataset$\\ $21 \times med\_dataset$\\ $21 \times med\_dataset$ \\ $21 \times med\_dataset$\end{tabular} \\
    \midrule
    \multirow{2}{*}{{D2}} & \multirow{2}{*}{\begin{tabular}[c]{@{}c@{}} \textbf{slow} \\ (Binary) \\ $prec \simeq 60\%$ \end{tabular}} & Lift &  
    \begin{tabular}[c]{@{}l@{}}$(s_5):\text{\texttt{\textcolor{blue}{GROUPBY}_stocks.gestion.modele.mvtrealise.refexterne}} > 0$\\ $(s_6):\text{serverName = ServerX} \wedge \text{systemI/O} > 50 $ \end{tabular}
    %\begin{tabular}[c]{@{}l@{}}$(s_5):\text{\texttt{\textcolor{blue}{GROUPBY}_stocks.gestion.modele.mvtrealise.refexterne}} > 0$\\ $(s_6):\text{serverName = ServerX\footnote[3]{Server names have been anonymized}} \wedge \text{systemI/O} > 50 $ \end{tabular}
    & \begin{tabular}[c]{@{}c@{}} 131 \\ 38   \end{tabular} & \begin{tabular}[l]{@{}l@{}} $prec = 100\%$ \\ $prec = 100\%$  \end{tabular} \\  \cmidrule(l){3-6} 
    & & WRAcc & \begin{tabular}[c]{@{}l@{}}$(s_7):\text{\texttt{\textcolor{blue}{WHERE}_stocks.gestion.modele.mvtrealise.etatsynchro}} > 0 \wedge \text{jdbcMax} < 200$ \\ $(s_8):\text{\texttt{\textcolor{blue}{WHERE}_stocks.gestion.modele.mvtrealise.auditinfo.datcre}} > 0 \wedge \text{dbVersion} = 2.3$ \\ $(s_9):$ manyActiveSessions $=$ Alarm  \end{tabular} & \begin{tabular}[c]{@{}c@{}} 20668 \\ 20675 \\ 44 \end{tabular} & \begin{tabular}[l]{@{}l@{}} $prec \simeq 99\%$ \\ $prec \simeq 99\%$ \\ $prec \simeq 93\%$  \end{tabular}  \\
    \midrule 
    \multirow{2}{*}{{D3}} & \multirow{2}{*}{\begin{tabular}[c]{@{}c@{}} \textbf{concurrence} \\ (Binary) \\ $prec \simeq 6\%$ \end{tabular}} & Lift & $(s_{10}):\text{\texttt{\textcolor{blue}{FROM}_.stocks.fichierbase.modele.produit}} > 0 \wedge \text{administrative} = 0.3$ & 8 & $prec = 100\%$ \\ \cmidrule(l){3-6} 
   % & & Binomial & $(s_{11}):\text{serverName} = \text{ServerY\footnote[3]{Server names have been anonymized}} \wedge \text{commit} > 0.7 \wedge \text{systemI/O} > 10.2$ & 51 & $prec \simeq 94\%$ \\ 
     & & Binomial & $(s_{11}):\text{serverName} = \text{ServerY} \wedge \text{commit} > 0.7 \wedge \text{systemI/O} > 10.2$ & 51 & $prec \simeq 94\%$ \\ 
    \midrule
    \multirow{2}{*}{{D4}} & \multirow{2}{*}{\begin{tabular}[c]{@{}c@{}} \textbf{blockedSess} \\ (Binary) \\ $prec \simeq 4\%$ \end{tabular}} & Lift & \begin{tabular}[c]{@{}l@{}} $(s_{12}):\text{\texttt{\textcolor{blue}{JOIN}_.commandesfactures.modele.histcdeligliv.applibudrist}} > 0 $ \\ $(s_{13}):\text{\texttt{\textcolor{blue}{WHERE}_ventes.commandesfactures.modele.cdeligliv.bonliv.datdepart}} > 0$ \end{tabular} & \begin{tabular}[c]{@{}c@{}} 7 \\ 9 \end{tabular} & \begin{tabular}[l]{@{}l@{}} $prec = 100\%$ \\ $prec \simeq 90\%$  \end{tabular} \\ \cmidrule(l){3-6}
     & & Binomial &  \begin{tabular}[l]{@{}l@{}} $(s_{14}):$ anomalyASH $=$ Critical \\ $(s_{15}):$ poolAlmostFull $=$ Info \end{tabular} & \begin{tabular}[l]{@{}l@{}} 151 \\ 124 \end{tabular} & \begin{tabular}[l]{@{}l@{}} $prec \simeq 85\%$ \\ $prec \simeq 99\%$  \end{tabular} \\
    \bottomrule
  \end{tabular} }
\end{table*}

\begin{figure}[h]
\centering
\subfloat[Subgroups distribution w.r.t time on D1 compared to overall data.]{
\includegraphics[width=0.4\textwidth]{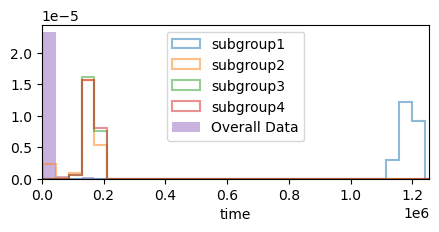}
\label{fig:subfig1}}
\qquad
\subfloat[Distribution of attributes in the table \texttt{cumulmultiple}.]{
\includegraphics[width=0.4\textwidth]{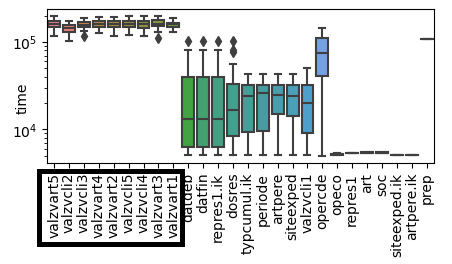}
\label{fig:subfig2}}
\caption{Statistical distributions of subgroups found on D1.}
\end{figure}

\smallbreak
\noindent \textbf{Results and Analysis.} Actionable and relevant subgroups have been identified in the different use-cases. In the dataset \textbf{D1}, we were interested in subgroups whose median execution time is significantly higher than the median of the dataset, while taking into account the subgroup size. We have chosen the measure \texttt{q_med} instead of the \texttt{Mean-test}, because we observed in particular for this example, that the mean is more sensitive to outliers. The subgroup $(s_1)$ which covers all queries that involve the attribute \texttt{auditinfo.etat} in the \texttt{\textcolor{blue}{WHERE}} clause has a very large median compared to the dataset, but only few objects. In Figure~\ref{fig:subfig1}, we show that its density distribution is too divergent and does not follow the usual distribution of original data. On the other hand, while the subgroup $(s_2)$ includes all the 451 queries executed on the \texttt{cumulmultiple} table characterized by a large median, the subgroups $(s_3)$ and $(s_4)$ are subsets of $(s_2)$ as they cover only its queries having the attributes \texttt{valzvcliX} and \texttt{valzvartX} respectively in their \texttt{\textcolor{blue}{WHERE}} clause. As shown In Figure~\ref{fig:subfig1}, 
the deviation of $(s_3)$ and $(s_4)$ from the overall distribution is stronger than the deviation of $(s_2)$,
%the distributions of $(s_3)$ and $(s_4)$ are much more deviant from the usual data distribution compared to $(s_1)$,
since they do not cover some slow queries present in $(s_2)$. To better understand this result, we have examined the \texttt{cumulmultiple} table by highlighting the distributions of its attributes in Figure~\ref{fig:subfig2}. We then confirm that mostly the attributes \texttt{valzvcliX} and \texttt{valzvartX} cause the $s_2$ to be identified. In \textbf{D2}, we discretize the attribute time so that queries with an execution time higher than 10 seconds are considered as slow queries. For each measure, we obtain interesting results that incorporate the extended features (e.g., alerts in $(s_9)$ and environment variables in $(s_7)$). For example, we found that all the queries on the \texttt{mvtrealise} table that are executed on ServerX\footnote{Server names have been anonymized} when the systemI/O is at least 50, last more than 10 seconds. Moreover, each time, the \texttt{refexterne} attribute is requested by the \texttt{\textcolor{blue}{GROUP BY}} clause, the query takes very long time to execute. Unlike the \texttt{lift} measure which relies only on the precision of the subgroup, \texttt{WRAcc} takes the subgroup size into account. The subgroups $(s_7)$ and $(s_8)$ are very exceptional because they cover more than 42\% of the queries while having an approximate precision of 99\% compared to an overall precision of 60\%  i.e., these subgroups contain more than 70\% of the slow queries. In \textbf{D3}, we aim to figure out the context in which queries encounter concurrency problems. The best results are achieved using the \texttt{lift} and \texttt{binomial} measures. Although the precision on the considered dataset is estimated to be only 6\%, we extracted subgroups with a precision that exceeds 94\%. The subgroup $(s_{11})$ alone constitutes 28\% of objects that characterize a concurrency issue. Finally, for \textbf{D4}, we want to extract relevant hypotheses that reveal the context in which the \textit{blocked sessions} alert is raised with blocking or critical level. This is generally due to the execution of a query which blocks a critical resource and puts new sessions on hold. We found that every time the table \texttt{applibudrist} is \textit{joined} with another table, the alert is triggered. Another possible reason may be a process that queries a table with missing indexes. This is where the subgroup $(s_{13})$ allows quickly to check if this assumption is true on the \texttt{datdepart} attribute. We also observed that the concerned alert is highly correlated to both the two alerts described by the subgroups $(s_{14})$ and $(s_{15})$. It is worth noting that the proposed SQL parser has been effectively useful, since it helped to contextualize interesting subgroups of queries. In fact, we notice through the experiments that the extracted subgroup patterns include different SQL clauses (e.g., \texttt{\textcolor{blue}{WHERE}}, \texttt{\textcolor{blue}{GROUP BY}}, etc.).

\subsection{Quantitative Analysis}
We study the time performance of both exhaustive and heuristic SD algorithms on the four datasets.
For each case, we set different values for the number of returned subgroups $k$, and the depth, i.e, the maximum number of selectors per pattern. Results are provided in Table~\ref{tab:sdruntime}.
The \texttt{limit} value refers to an execution time that exceeds $1,000$ seconds. Beam-Search was able to finish in less than $274$ seconds for all configurations, while Depth-First exceeded $1,000$ seconds in many cases. 
Exceptionally in \textbf{D3}, Depth-First took less time than Beam-Search. This may be due to the relatively small number of features compared to other datasets. These results also show the impact of parameters, i.e., the number of returned patterns and the depth. In fact, the higher they are, the longer is the execution time.
In our qualitative experiments, we use Beam-Search with a beam width of 50 for D1 and D4, while we exploit results from Depth-First on D2 and D3. 

\begin{table}
\centering
\caption{Execution time (in seconds) of SD algorithms.}
\renewcommand{\arraystretch}{1} 
\label{tab:sdruntime}
\scalebox{0.77}{
\begin{tabular}{@{}c|c|c|c|c|c|c|c|c@{}}
\textbf{Algo}        & \multicolumn{4}{c|}{\textbf{Beam-Search} (heuristic)}         & \multicolumn{4}{c}{\textbf{Depth-First} (exhaustive)}   \\ \midrule
\textbf{\# patterns (k)} & \multicolumn{2}{c|}{10} & \multicolumn{2}{c|}{50} & \multicolumn{2}{c|}{10} & \multicolumn{2}{c}{50} \\ \midrule
\textbf{depth} & 2     & 3     & 2      & 3      & 2      & 3     & 2     & 3     \\ \midrule
\textbf{D1} & 20.17 & 90.39 & 113.71 & 274.60 & \texttt{limit}  & \texttt{limit} & \texttt{limit} & \texttt{limit} \\ \midrule
\textbf{D2} & 5.35  & 5.40  & 28.44  & 45.04  & 440.01 & \texttt{limit} & 458.2 & \texttt{limit} \\ \midrule
\textbf{D3} & 0.75  & 0.83  & 3.94   & 4.27   & 0.18   & 0.50  & 0.53  & 0.59  \\ \midrule
\textbf{D4} & 10.73 & 10.98 & 56.30  & 62.35  & \texttt{limit}  & \texttt{limit} & \texttt{limit} & \texttt{limit}
\end{tabular}}
\end{table}

\subsection{Interactive Subgroup Discovery\label{subsec:interactive}}
\begin{figure*}
\centering
\includegraphics[width=0.93\textwidth]{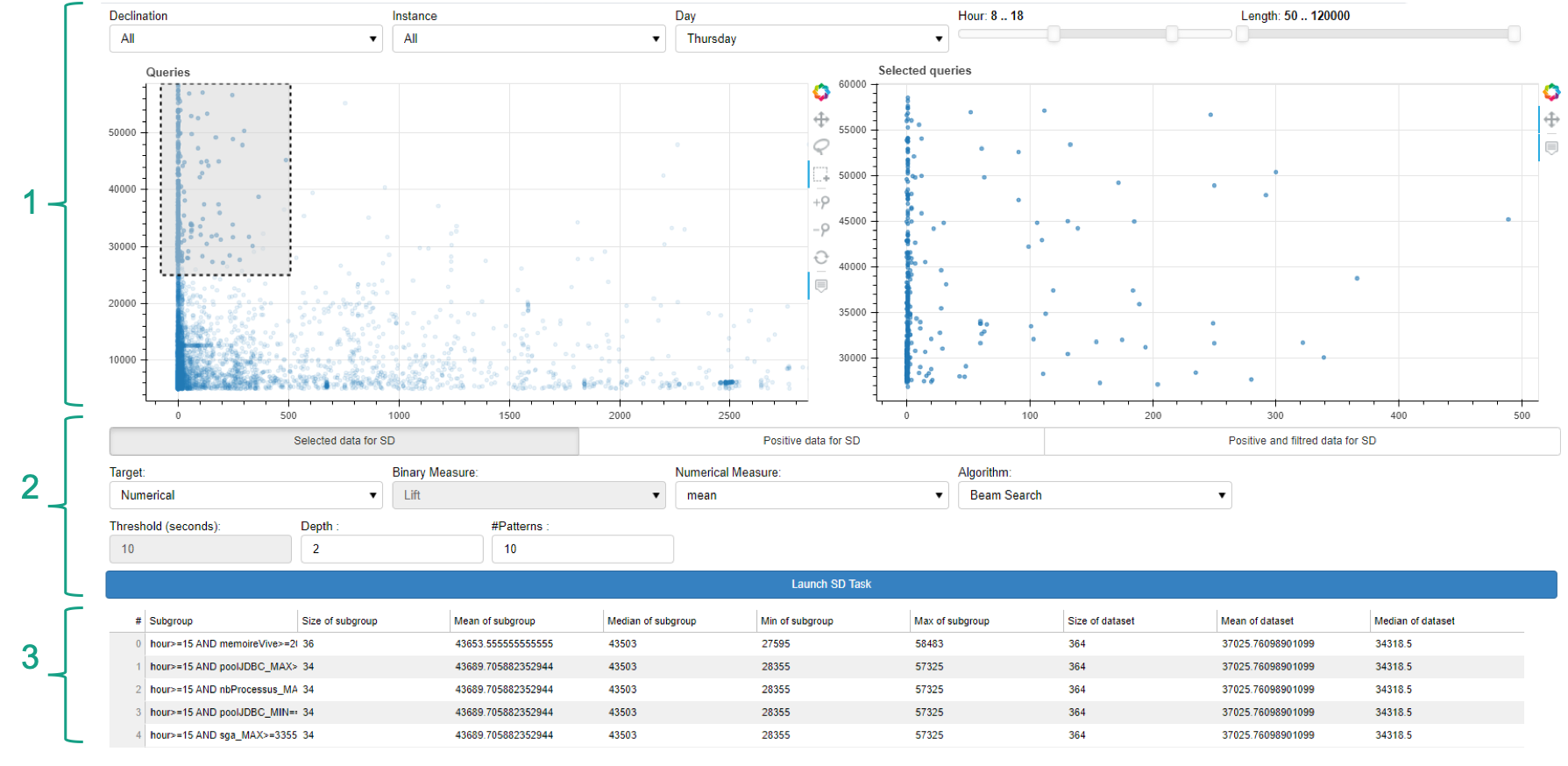}% &
\caption{\label{fig:interacttool} Main sections of the interactive SD tool: (1) dataset properties, (2) search strategy, and (3) results.}
\end{figure*}%
Figure~\ref{fig:interacttool} illustrates our interactive visualisation: it can manage different data types, both for input features as well as the target, including nominal and numerical attributes. It also provides a range of interestingness measures and algorithms. Its main window consists of 3 important panels: (1) the  dataset properties, (2) the search strategy, and (3) the results. 

\smallbreak
\noindent \textbf{Dataset panel.} It allows the user to select a subset of data of interest thanks to data points selection where queries are plotted, e.g., w.r.t. execution times and row counts, but also filters on query properties. The graph on the right simply rescale the selection made on the left graph.

\smallbreak
\noindent \textbf{Search strategy panel.} This panel enables the configuration of the mining task. First, the user needs to define the target. There are two possibilities: (1) choose a specific attribute as target, or (2) graphically create a binary target by associating its positive class to the data subset selected in the right graph, and the remaining data of the left graph as negative class. After that, she can specify the interestingness as well as the mining algorithm. Finally, it remains to set the desired number of returned subgroups, and the maximum depth of pattern-rules.

\smallbreak
\noindent \textbf{Results.} Once the mining task is executed, this panel shows the identified subgroups. For each, it displays the corresponding pattern along with relevant statistics such as the subgroup size, the median, etc.
\section{Related Work}\label{sec:relatedwork}

For decades, extracting interesting patterns from query workloads has been of great importance in database research. A variety of related methods have been proposed to perform specific tasks on workloads. 
The use case most closely related to our solution is performance analysis~\cite{chaudhuri2002compressing,deep2020comprehensive,jain2018query2vec}. However, most of these approaches used clustering-based methods which are not practical to identify subsets of data that specifically discriminate a property of interest. On the other hand, Several major commercial database systems have developed tools to automate this task such as query planner and optimizer; Microsoft SQL Server has included the index selection feature as part of its Tuning Advisor since SQL Server 2000 \cite{DBLP:conf/sigmod/AgrawalCKMNS05}. Even if these tools are widely used by DBAs, they remain specific and non-generic tools as they are limited to certain features. Indeed, using query optimizer for example, requires digging into individual cases to figure out the issue in each query separately, while a Subgroup Discovery approach aims to identify issues for a \textit{subset} of queries sharing some specific properties w.r.t any user-defined target. Moreover, it can be argued that Subgroup Discovery may be used to assist query planner with very specific cases i.e., providing interesting cases to be investigated with query optimizer. We are the first to address the challenging task of adapting Subgroup Discovery for a complex and generic Workload Analysis problem. 
In what follows, we describe in more details the different Workload analysis tasks that have been studied in the literature. 

\smallbreak
\noindent\textbf{Performance Optimization.} Database system performance can be tuned by recommending the appropriate set of indexes to speed up query processing. However the complexity of index selection grows quadratically with the workload size \cite{deep2020comprehensive}. Therefore, several approaches \cite{deep2020comprehensive,chaudhuri2002compressing,jain2018query2vec, chaudhuri2003primitives} tackle this challenge by finding a compressed workload that is highly representative i.e., a smaller substitute workload that has similar performance characteristics as the original workload.  
This compression problem is NP-Hard \cite{chaudhuri2002compressing}. Thus, existing approaches have used a variety of heuristic techniques ranging from random sampling \cite{chaudhuri2002compressing} and clustering \cite{chaudhuri2003primitives} to the use of sophisticated Machine Learning models \cite{jain2018query2vec}. For instance, \cite{chaudhuri2002compressing,chaudhuri2003primitives} 
use a distance function that measures the difference between pairs of SQL statements, with respect to the workload-driven task. Then, they propose multiple summarization techniques including K-Medoids, random sampling and all pairs greedy algorithm. More recently, \cite{xie2018query} propose query structure based clustering algorithms that rely only on the syntactic information of the query. Query2Vec \cite{jain2018query2vec} cluster queries based on representations computed using several NLP approaches.

\smallbreak
\noindent \textbf{Insider threats identification.} \cite{kul2016ettu} propose a semi-supervised approach to analyse database access patterns. It starts by clustering SQL queries using a similarity function that is defined over query structures. Some of these clusters are labeled by experts as potential security threats. Then, these labels are used to generate patterns that enable the automatic classification of remaining clusters.

\smallbreak
\noindent \textbf{Query Recommendation.} This task aims to assist non-expert users by providing them with personalized SQL query recommendations that correspond to their information needs. \cite{akbarnejad2010sql} exploit a collaborative filtering paradigm where users with similar querying behavior are assumed to be interested in retrieving similar data. \cite{aligon2014similarity} introduce an order-sensitive model to compare OLAP user sessions where the order of queries within a session influences the similarity of sessions. ExplIQuE~\cite{DBLP:conf/cikm/GuillyPSI19} uses clustering and decision trees to extend a given query,  by suggesting a set of possible selection predicates to add to the query, that aim at dividing the initial answer set to identify interesting exploration zones.

\smallbreak
\noindent \textbf{Finding user interest.} This problem aims to track the user's historical querying behavior to seek for her interest. In \cite{arzamasova2019usefulness}, the authors propose a query similarity metric based on the notion of the so-called access area. This area captures the part of the data space that the user is mostly interested in. Another related work \cite{chen2007addressing} compares queries based on returned results, then cluster the data to help users locate interesting results.

\smallbreak
\noindent \textbf{Antipatterns Detection.} This problem consists in extracting patterns that generally lead to unnecessary SQL statements which may have a negative effect on performance, or introduce bias on any subsequent workload analysis~\cite{arzamasova2017cleaning}. In~\cite{eessaar2015query}, the proposed method analyzes metadata tables to detect \textit{design} antipatterns that reflect errors in the database schema. Whereas \cite{arzamasova2017cleaning,chen2014detecting} are interested in antipatterns related to performance degradation. \cite{chen2014detecting} use static code analysis and rule-based approach, while in \cite{arzamasova2017cleaning}, the final goal behind cleaning query logs is to simplify the identification of interests of database users.

%\smallbreak
%\noindent \textbf{Automated Grading.} Automated Grading is a specific task in the educational context that aims to automatically check whether the student queries are correct or not. For grading student assignments, it is usually not sufficient to just check the correctness of the query, but also being able to attribute a partial score if the query is incorrect based on how close the query is to being correct \cite{chandra2016partial}. In order to ensure whether a given student’s SQL statement is a valid answer to a given problem, most tools \cite{kleiner2013automated,brusilovsky2010learning,de2006students}, compare the outcome of the statement with one or many reference solutions. SQLator \cite{sadiq2004sqlator} in contrast uses various heuristics to detect similarity between the solution and the reference statement. Following this idea, it is possible that correct solutions cannot be detected. An intermediate approach is proposed later by \cite{dollinger2010sql}, who uses result comparison first to check correctness, and semantic comparison by heuristic reformulation rules to give hints to the student.  

\smallbreak
\noindent \textbf{Visual Analysis.} Makiyama et al. \cite{makiyama2015text} use SOM (Self Organizing Map) as a visualization tool due to its quantization and projection properties. The provided visualization gives an idea of the overall shape of the data and helps to detect possible cluster structures in the SQL workload. QueryScope \cite{hu2008queryscope} aims to find better tuning opportunities by helping users identify shared patterns between queries while providing a variety of viewing options so that a user can focus on query relevant aspects.

\section{Conclusion and Perspectives}\label{sec:conclusion}
Mining patterns in SQL workloads helps DBAs discovering subgroups of queries sharing some properties and strongly discriminating either a nominal target or a metrics property. We developed a methodology based on Subgroup Discovery which can be tuned in terms of descriptive and target attributes, mining algorithms and pattern quality measures. We proposed a visualisation tool for helping practitioners to make subgroup discovery possible with an interactive platform. We empirically showed how it can elicit hypotheses of interest from queries run on hundreds of databases. We are currently working on integrating our approach in a large scale supervision framework for daily preventive maintenance for our DBA team. Subgroup discovery can be extended in many ways to provide better results. First, we realized through experiments that we often need to consider multiple targets, for example to identify patterns of queries which return few rows while having high running times. Second, although we consider a rich pattern language in comparison to other approaches, we can exploit the syntactic tree structure of the queries, to mine tree patterns, more expressive than conjunctions of SQL clauses. Third, an effort is needed to produce more qualitative subgroup sets with more diversity (data cover) and less redundancy \cite{DBLP:journals/datamine/BoscBRK18}, and directly considering a quality measure on the subgroup set, turning the \textit{top-k mining problem} into \textit{subgroup set mining} \cite{DBLP:conf/dsaa/BelfodilBBLRKP19}. Finally, subjectivness and practitioner preferences should be considered by the mining algorithms through an interactive discovery process. These are actually current challenges in the field of subgroup discovery.

\bibliographystyle{IEEEtran}
\bibliography{IEEEabrv,references}

\appendix

\subsection{Comparison between interestingness measures}

\begin{figure}[htb]
\centering
\subfloat[\texttt{mean}.]{
\includegraphics[width=0.4\textwidth]{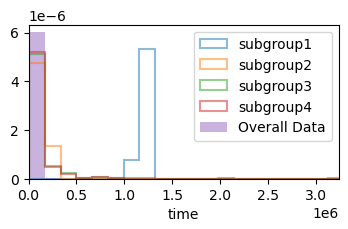}
\label{fig:subfig1dist}}
\qquad
\subfloat[\texttt{median}.]{
\includegraphics[width=0.4\textwidth]{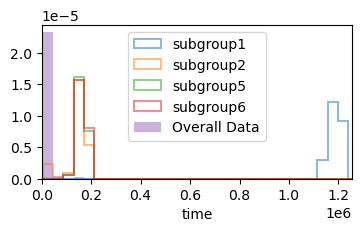}
\label{fig:subfig2dist}}
\qquad
\subfloat[T-score.]{
\includegraphics[width=0.4\textwidth]{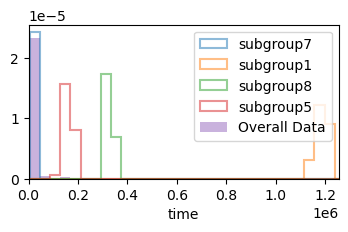}
\label{fig:subfig3dist}}
\caption{\label{fig:suppdist}Statistical distributions of subgroups found on D1 for different measures.}

\end{figure}

In this section, we aim at comparing the resulting subgroup patterns when using different measures of interest. In particular, our study is conducted on the two examples D1 and D2 defined previously in the Section~\ref{sec:evaluation}. In the first example D1, we chose the \texttt{median} instead of the \texttt{\texttt{mean}} because the \texttt{mean} is very sensitive to outliers. The results found for each measure are presented in Table~\ref{tab:suppresults}. Note that we can obtain similar subgroup patterns for different interestingness measures as shown in Figure~\ref{fig:suppdist}. For instance, the subgroup $(s_1)$ is always identified, regardless of the measure of interest used for the subgroup discovery approach. Different from the \texttt{median}-based approach, the two subgroups $(s_3)$ and $(s_4)$ are identified as interesting subgroups w.r.t. to the \texttt{mean} measure. However, by analysing their distributions, we do not observe a significant divergence from the original distribution compared to $(s_{5})$ and $(s_{6})$. Indeed, the subgroups $(s_3)$ and $(s_4)$ contain some slow queries but still also contain many queries that execute quickly. On the other hand, the \texttt{T-score} based approach incorporates the mean and the standard deviation of the target value to reflect the cohesion of the subgroup. Except the subgroup $(s_3)$ that has a \texttt{support} of 45, identified subgroups are small in size. This does not match the assumption on this measure which requires the subgroup to contain at least 30 objects. In the second example, we show that frequent patterns that rely only on the \texttt{support} measure, are not always interesting. For example, the subgroup $(s_{10})$ contains slow queries that takes more than 10 seconds to execute, but its precision is poor compared to \texttt{Lift} and \texttt{WRAcc} measures. The \texttt{Lift} measure evaluates the subgroups based only on the precision. Usually with lift, we end up with small subgroups but with high precision. In contrast, \texttt{WRAcc} and \texttt{Binomial} measures depend on the subgroup size. This means that one subgroup can be prioritized over another subgroup that has more precision but contains far fewer objects.      

\begin{table*}
\caption{Subgroup Discovery Results with different measures}
\renewcommand{\arraystretch}{1} 
\label{tab:suppresults}
\centering
\scalebox{0.82}{
\begin{tabular}{c|c|c|l|c|c}
\toprule 
\textbf{ID}         & \textbf{Target }  & \textbf{Measure}  & \multicolumn{1}{c|}{\textbf{Subgroup patterns}} & \textbf{Size} & \textbf{Quality} \\ \midrule
\multirow{3}{*}{D1} & \multirow{3}{*}{\begin{tabular}[l]{@{}c@{}} \textbf{time} \\ (Numerical) \end{tabular}}    & \texttt{mean}     & \begin{tabular}[l]{@{}l@{}} $(s_1):\text{\texttt{\textcolor{blue}{WHERE}_stocks.gestion.modele.lot.prod.ref.auditinfo.etat}} > 0$ \\ $(s_2):\text{\texttt{\textcolor{blue}{FROM}_ventes.cumuls.modele.cumulmultiple}} > 0$\\ $(s_3):\text{\texttt{\textcolor{blue}{WHERE}_stocks.gestion.modele.mvtrealise.flagaepurer}} > 0$ \\ $(s_4):\text{\texttt{\textcolor{blue}{SELECT}_stocks.gestion.modele.mvtrealise.ik}} > 0$\end{tabular} & \begin{tabular}[c]{@{}c@{}}8\\ 451\\ 602 \\ 719\end{tabular}   &   \begin{tabular}[l]{@{}l@{}}$78 \times \texttt{mean}\_dataset$\\ $9 \times \texttt{mean}\_dataset$\\ $6 \times \texttt{mean}\_dataset$ \\ $6 \times \texttt{mean}\_dataset$\end{tabular}      \\ \cmidrule(l){3-6} 
                    &                          & \texttt{median}   & \begin{tabular}[l]{@{}l@{}} $(s_5):\text{\texttt{\textcolor{blue}{WHERE}_ventes.cumuls.modele.cumulmultiple.valzvcli\textbf{X}}} > 0$ \\ $(s_6):\text{\texttt{\textcolor{blue}{WHERE}_.ventes.cumuls.modele.cumulmultiple.valzvart\textbf{X}}} > 0$\end{tabular}   
                    & \begin{tabular}[c]{@{}c@{}} 45 \\ 45\end{tabular} & \begin{tabular}[l]{@{}l@{}}$21 \times med\_dataset$ \\ $21 \times med\_dataset$\end{tabular}        \\ \cmidrule(l){3-6} 
                    &                          & T-score  &  \begin{tabular}[l]{@{}l@{}} $(s_7):\text{\texttt{\textcolor{blue}{WHERE}_stocks.achats.cadencier_fournisseur.modele.cadencier.mat.art.ik}} > 0$ \\ $(s_8):\text{\texttt{\textcolor{blue}{WHERE}_achats.fournisseurs.modele.fourlivperiodereg.datfin}} > 0$\end{tabular}                 & \begin{tabular}[c]{@{}c@{}} 8 \\ 2 \end{tabular} & \begin{tabular}[l]{@{}l@{}} - \\ - \end{tabular}        \\ \midrule
\multirow{4}{*}{D2} & \multirow{4}{*}{\begin{tabular}[c]{@{}c@{}} \textbf{slow} \\ (Binary) \\ $prec \simeq 60\%$ \end{tabular}} 

& Support  &  \begin{tabular}[l]{@{}l@{}} $(s_9):\text{\texttt{\textcolor{blue}{WHERE}_stocks.gestion.modele.mvtrealise.lot.ik}} > 0$ \\ $(s_{10}):\text{dbCursorsMax} < 2000$\end{tabular} &  \begin{tabular}[c]{@{}c@{}} 29827 \\ 30548\end{tabular}    &  \begin{tabular}[l]{@{}l@{}} $prec \simeq 65\%$ \\ $prec \simeq 65\% $  \end{tabular}       \\ \cmidrule(l){3-6} 

                    &                          & Lift     &     \begin{tabular}[c]{@{}l@{}}$(s_{11}):\text{\texttt{\textcolor{blue}{GROUPBY}_stocks.gestion.modele.mvtrealise.refexterne}} > 0$\\ $(s_{12}):\text{serverName = serverX} \wedge \text{systemI/O} > 50 $ \end{tabular} & \begin{tabular}[c]{@{}c@{}} 131 \\ 38   \end{tabular} & \begin{tabular}[l]{@{}l@{}} $prec = 100\%$ \\ $prec = 100\%$  \end{tabular}       \\ \cmidrule(l){3-6} 
                    
                    &                          & \begin{tabular}[c]{@{}c@{}} WRAcc / \\ Binomial \end{tabular}   &   \begin{tabular}[c]{@{}l@{}}$(s_{13}):\text{\texttt{\textcolor{blue}{WHERE}_stocks.gestion.modele.mvtrealise.etatsynchro}} > 0 \wedge \text{jdbcMax} < 200$ \\ $(s_{14}):\text{\texttt{\textcolor{blue}{WHERE}_stocks.gestion.modele.mvtrealise.auditinfo.datcre}} > 0 \wedge \text{dbVersion} = 2.3$  \end{tabular} & \begin{tabular}[c]{@{}c@{}} 20668 \\ 20675  \end{tabular} & \begin{tabular}[l]{@{}l@{}} $prec \simeq 99\%$ \\ $prec \simeq 99\%$   \end{tabular}        \\  \bottomrule

\end{tabular}}[h]
\end{table*}

\subsection{Effectiveness of the postprocessing phase}

\begin{figure*}[ht]
\centering
\subfloat[Hierarchical Clustering of all patterns.]{
\includegraphics[width=0.46\textwidth]{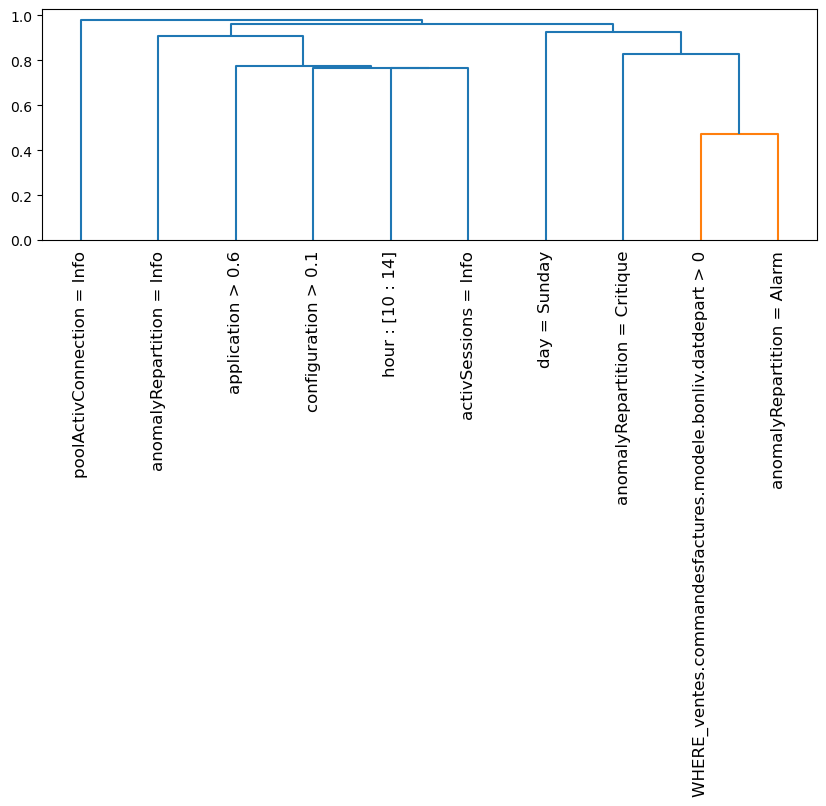}
\label{fig2:subfig1clust}}
%\qquad
\subfloat[Truncated Hierarchical clustering.]{
\includegraphics[width=0.46\textwidth]{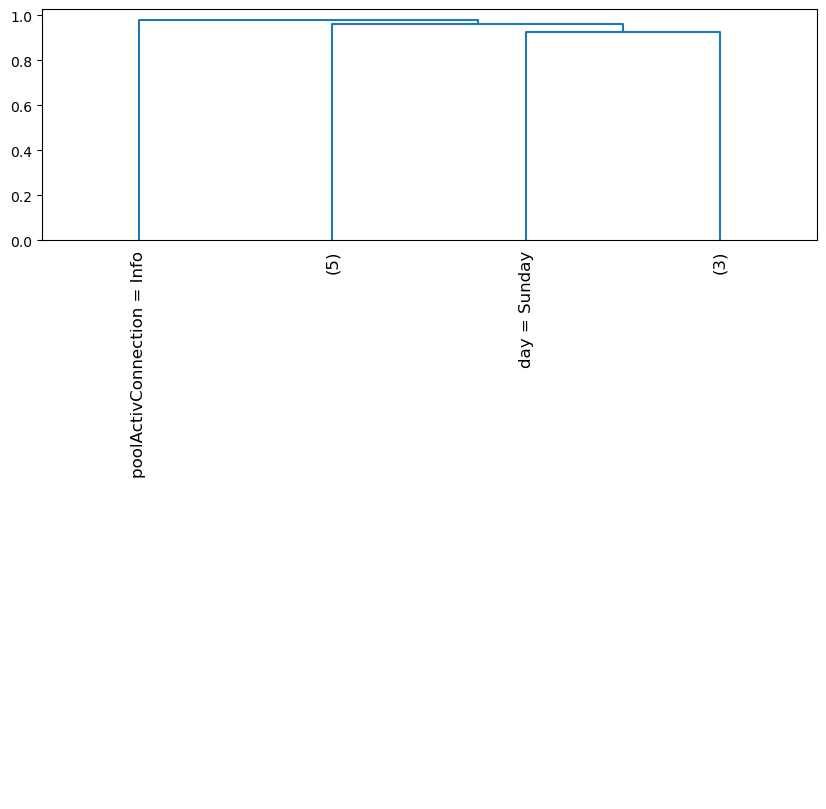}
\label{fig2:subfig2clust}}
\caption{\label{fig:supppostproc}\textbf{Post processing.} Hierarchical clustering based on \texttt{Jaccard} distance }

\end{figure*}

When selecting interesting subgroups based only on the measure of interest, we may end up with redundant subgroups. In this section, we show how the postprocessing phase is useful in providing the user with interesting but diverse patterns. We use the Agglomerative Hierarchical clustering on example D4 to extract the most representative subgroups that correlates with the alert \textit{blockedSessions}. In the example shown in Figure~\ref{fig:supppostproc}, we initially extract the top-10 subgroups using the \texttt{Binomial} Measure. We then perform the clustering on the 10 found subgroups based on the \texttt{Jaccard distance}. Afterwards, we chose a partition of (4) different subgroups by truncating the tree at the distance (0.91). This means that we allow only subgroups having at most 0.09 of similarity between them. We end up with 2 subgroups patterns and two clusters that contains 5 and 3 subgroups respectively. For each cluster we choose the best subgroup w.r.t to its measure of interest, thus we end up with the two patterns that we displayed in Section~\ref{sec:evaluation}.

\end{document}